\documentclass{amia}
\usepackage{graphicx}
\usepackage[labelfont=bf]{caption}
\usepackage[superscript,nomove]{cite}
\usepackage{color}
\usepackage{subfigure}
\usepackage{multirow}
\usepackage{wrapfig}
\usepackage{hyperref}
\begin{document}
\renewcommand{\thefootnote}{\roman{footnote}}

\title{Understanding Heart-Failure Patients EHR Clinical Features via SHAP Interpretation of Tree-Based Machine Learning Model Predictions}

\author{Shuyu Lu, MS$^{1}$, Ruoyu Chen, PhD$^2$, Wei Wei,PhD$^{3,4}$ , Xinghua Lu MD, PhD$^1$  }

\institutes{
    $^1$Dept. Biomedical Informatics, University of Pittsburgh, Pittsburgh, PA, USA;\\
    $^2$Computer School, Beijing Information Science \& Technology University, Beijing, China\\
    $^3$University of Pittsburgh Medical Center, Pittsburgh, PA, USA;\\ $^4$Astrata Solutions, Pittsburgh, PA, USA
}

\maketitle

\noindent{\bf Abstract}

\textit{Heart failure (HF) is a major cause of mortality. Accurately monitoring HF progress and adjust therapies are critical for improving patient outcomes. An experienced cardiologist can make accurate HF stage diagnoses based on combination of symptoms, signs, and lab results from the electronic health records (EHR) of a patient, without directly measuring heart function. We examined whether machine learning models, more specifically the XGBoost model, can accurately predict patient stage based on EHR, and we further applied the SHapley Additive exPlanations (SHAP) framework to identify informative features and their interpretations. Our results indicate that based on structured data from EHR, our models could predict patients’ ejection fraction (EF) scores with moderate accuracy. SHAP analyses identified informative features and revealed potential clinical subtypes of HF. Our findings provide insights on how to design computing systems to accurately monitor disease progression of HF patients through continuously mining patients’ EHR data.} 

\section*{Introduction}
Heart failure, also commonly referred to as congestive cardiac failure, is a clinical syndrome when the heart is unable to pump sufficiently to maintain blood flow to meet the body’s needs. According to CDC, HF is one of the most prevalent diseases and with highest mortality rate \cite{ref1} within the US, with the estimating that 6.2 million adults are affected\cite{ref1}. Accurately monitoring disease progression and timely adjusting regimen to prevent or slow progression would improve HF patients' life expectancy and quality. With prevalent availability of EHR, continuously mining EHR by computation agents to accurately detect and report disease progression will likely become a component of long term care of HF patients.  In this study, we investigated the feasibility of using computational agents to assess disease stage and characteristic clinical features that provide information of disease progression. 

\begin{wraptable}{r}{0.62\textwidth}
    \centering
    \begin{tabular}{ccc}
        \hline
        \textbf{EF Score}&  \textbf{Pumping Ability of the Heart} & \textbf{Level of Heart Failure}\\
        \hline
        50\%-70\% & Normal & No HF/HFpEF \\
        40\%-50\% & Slightly below Normal & Slight Symptoms of HF\\
        35\%-40\% & Moderately below normal & Mild HFrEF\\
        $<$35\% & Severely below normal & Severe HFrEF \\
        \hline
    \end{tabular}
    \caption{Relation between EF score and HF\cite{ref3}}
    \label{EF HF}
\end{wraptable}

An objective measurement of heart function is the EF of left ventricle of heart. Broadly classified, there are 2 types of heart failure: heart failure with reduced ejection fraction (HFrEF) and heart failure with preserved ejection fraction (HFpEF). The difference lies in whether the systolic or diastolic capabilities of the left ventricle is affected\cite{ref2}, which is primarily measured by the left ventricle ejection fraction (LVEF or EF) score. Basically, the relation between EF score and heart failure is shown in \textbf{Table 1}. According to 2016 European Society of Cardiology (ESC) Clinical Practice Guidelines\cite{ref3}, an EF score less than 40\% typically suggests HFrEF, while an EF score greater than 50\% is often a sign of HFpEF. As an exact numerical value, therefore, compared to other symptoms of heart failure, EF score is a desirable indicator when applying machine learning methods to predict a patient’s heart failure conditions. Currently, EF score is usually measured using echocardiogram during inpatient/outpatient hospital visits, but the frequency of echocardiogram can be far and between in comparison to recordings of other clinical data.  Furthermore, encounters of a HF patient with a health system are multi-facet and occurs in multiple settings, e.g., outpatient visit, nurse call, pharmacy visits, etc. All these encounter are usually recorded in EHR of a comprehensive health system like the UPMC. Therefore, EHR is an ideal resource for exploring indicators for diagnoses and outcomes. The continuously mining of EHR to detect disease progress would be of high clinical value in future.  

Machine learning models have been increasingly applied to explore medical scenarios related to heart failure \cite{ref6,ref7,ref8}.  In general, interpretable models are preferred over "black box" models in clinical settings. Among modern machine learning models, tree-based models, e.g., the XGBoost\cite{ref17} is often favored in clinical settings,  because they are easy to interpret, capable of handling missing values, and solves overfitting and underfitting problems effectively with the help of regularization methods. But most of previous studies concentrate on  model performance or feature importance\cite{ref6,ref7,ref8}, with little attention to further interpret the predictions with interpretable methods. That is, a human user not only is interested to know what features are informative with respect to a prediction task, but also would like to know how to interpret an observed value of a feature with respect to the prediction task. To this end, machine learning methods have been developed to identify informative features and their interpretations in a model-agnostic fashion \cite{ref5,ref9,ref10,ref11,ref12,ref13}. For example, a model-agnostic interpretation methods, such as SHAP \cite{ref4}, can take a dataset and different prediction models as inputs, apply the models to the data, and discover the characteristics of data features in each prediction model (thus model-agnostic) different prediction models.  

In this paper, we investigated the utility of XGBoost model in predicting heart failure stages, which is represented by EF scores based on structured EHR data. We evaluated informative features and investigated their interpretability and characteristics SHAP framework. Finally, based on the characteristics of features and their values, we applied unsupervised cluster learning to discover subtypes (subpopulations) among HF patients. To our best knowledge, few studies attempt to address the same questions as reported here, and we anticipate that our approach lays a foundation for future development of computation agents capable of monitoring HF patient disease progress through continuously mining the EHR data of health systems.

\section*{Materials and Methods}
\textit{Data Collection and Preprocessing}

All the EHR data were obtained from UPMC during 2014 to 2019 and contains clinical and medical conditions of the patients who have been attacked by HF before with ICD9/ICD10 codes (I428.* and I50.*) indicating HF diagnosis. The original dataset consisted of 9 different CSV files that were  extracted from UPMC EHR system, including demographics(DEMO\_*), vitals(VL\_*), labs(LB\_*), medical dispenses(MD\_*), medical fills(MF\_*), medical orders(MO\_*), order result(OR\_*), problem list(PL\_*), and diagnoses(DI\_*). Each patient has a unique ID, and each file is a collection of data from 60,835 unique patients and in total more than 20,000,000 encounters. In order to build a profile for each patient, we cleaned the raw CSV files based on rules, extracted each patient’s record from the CSV files and finally aggregated the information. The workflow is illustrated in \textbf{Figure 1}:

\begin{wrapfigure}{l}{0.6\textwidth}
\centering
\includegraphics[width=0.6\textwidth]{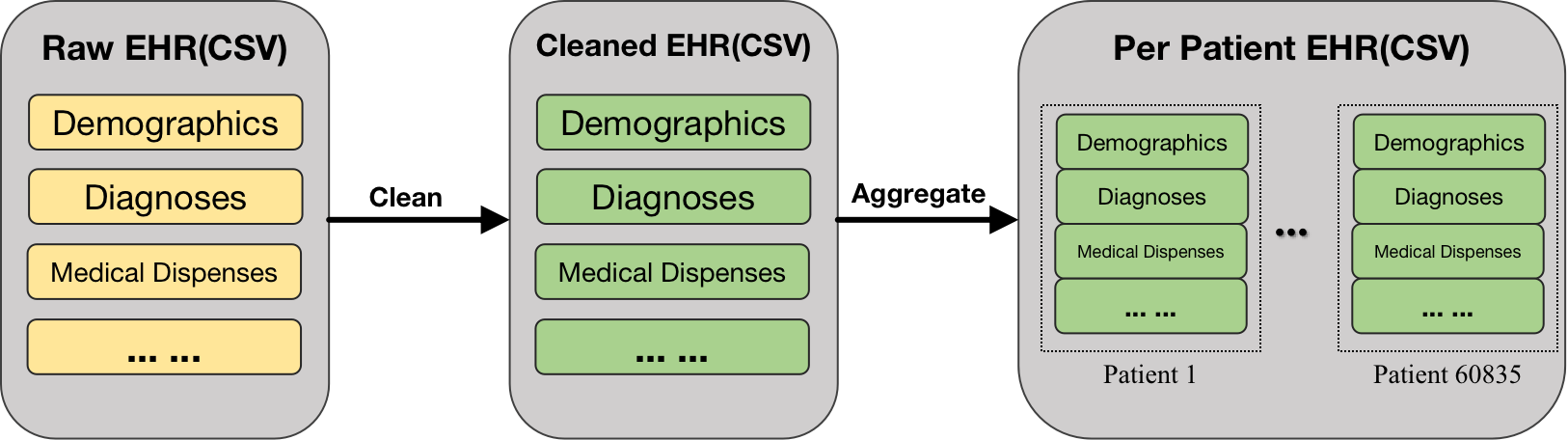}
\caption{Data clean \& aggregate workflow}
\label{fig1}
\end{wrapfigure}

The rules for data processing were:

\hangindent 2em
\hangafter=0
1) For medical fills, medical orders, medical dispenses, problem lists and diagnoses, only keep the drug and disease names that appear over 10,000 times in the dataset as valid features; 

\hangindent 2em
\hangafter=0
2) For numerical features like age, BMI, and blood pressure, in order to exclude outliers as much as possible, we normalize the values and only keep the ones between 1\% and 99\% percentile, and values outside this ranged are then set to the value of 1 percentile (MIN) or 99 percentile (MAX) ; 

\hangindent 2em
\hangafter=0
3) The medical fills, medical orders, and medical dispenses are mixture of the National Drug Code (NDC) and the Anatomical Therapeutic Chemical (ATC) code. The problem lists and diagnoses are mixture of the ICD-9 and ICD-10 code.  For the sake of consistency, all drugs NDC codes are mapped to ATC codes, and all   ICD-9 codes are mapped to ICD-10 codes with the help of Apache Lucene\cite{ref16}.

After data cleaning, 1894 features from the 9 categories were adopted and the values were stored tables. Details are available in
\textit{\href{https://github.com/Frank-LSY/XGB-SHAP-EHR-EF}{Feature Table}$\footnote{https://github.com/Frank-LSY/XGB-SHAP-EHR-EF}$}.

\textit{Training data}

To train and test models, we selected the patients with a valid EF score, which is a predominant indicator for HF diagnosis. The EF score also serves as the label in our model because it is an objective measure and easy to interpret, in addition it can be smoothly fit into the SHAP framework.

The EF scores were directly obtained from patients’ echocardiogram reports. We extract the rest of structure EHR data of a patient within 45 days of EF measure, and we processed the EHR to derive a feature vector matching an EF measurement. Since each patient can have multiple EF measurement across time, and a patients disease may progress through such time, we treat EF measurements and match features from a patient that were separated by more than 180 days as "independent" cases. Finally, we obtained 130,727 cases. We than split the dataset to 70\% for training, 20\% for validation and 10\% for test.

\begin{wraptable}{r}{0.4\textwidth}
\centering
  \begin{tabular}{|l|l|}
  \hline
    \textbf{Parameter}    & \textbf{Value} \\ \hline
    n\_estimators (\# of trees)   &   100  \\ \hline
    max\_depth   &   3 \\ \hline
    eta (learning\_rate) & 0.35 \\ \hline
    min\_child\_weight & 1 \\ \hline
    col\_sample\_by\_tree  & 1 \\ \hline
    col\_sample\_by\_level & 1 \\ \hline
    subsample & 0.85 \\ \hline
    reg\_alpha (L1 regularization) & 0 \\ \hline
    reg\_lambda (L2 regularization) & 0.5 \\ \hline
    gamma & 0 \\ \hline
    num\_boost\_round (\# of boosting iterations) & 500 \\ \hline
  \end{tabular}
  \caption{XGBoost fine-tuned parameters}
\end{wraptable}

\textit{XGBoost Model for Regression}

We applied XGBoost\cite{ref17} for our prediction task. XGBoost is optimized distributed gradient boosting library and has been successfully applied in medical studies\cite{ref18,ref19}.The idea behind the boosting algorithms is to combine weak classifiers together to form a powerful one. XGBoost is a boosting tree model that combines many CART regression tree models.

The regression tree model is trying to predict the original EF score with structured tabular EHR data. We tuned several hyper-parameters with the evaluation of 5-folder cross validation for the XGBoost model before reaching the final model. The parameters are tuned with sklearn.model\_selection.GridSearchCV\cite{ref20} package, coordinate descent\cite{ref21} as the strategy. The final tuned parameters are listed in \textbf{Table 2}:

\textit{SHAP for Model-Agnostic Interpretation}

SHAP stands for SHapley Additive exPlanations. It is a game theoretic approach to explain the output of ML model \cite{ref4}. It connects optimal credit allocation with local explanations using the classic Shapley values \cite{ref22} from game theory and their related extensions. SHAP would assign each sample, each feature a unique SHAP value for the particular prediction. The SHAP value represents the deviation from the average predicted value for each case prediction brought by each feature. For our work, we applied SHAP model to firstly generate SHAP value for all our test dataset cases, then illustrate the SHAP summary plot and SHAP scatter plot using publicly available SHAP  API \cite{ref23} for global understanding our dataset.

\textit{T-SNE clustering with SHAP values}

T-SNE (t-distributed Stochastic Neighbor Embedding) is a visualization machine learning algorithm based on stochastic neighbor embedding\cite{ref24}. In particular, it models each high-dimensional object with a 2D or 3D point, with similar data points placed as nearby points and dissimilar objects modeled with high probability by distant points.

For our work, after generating SHAP values for the dataset, more specifically each feature of a subject is assigned with a SHAP value, we clustered data points by their SHAP values associated with each feature (instead of in original feature space), and we examined  whether this approach could reveal different subtypes of HF patients by inspecting the clusters. We applied t-SNE algorithm to map the 1894 features to 2D for visualizing the inner relations.

\textit{Model implementing details}

The experiments were conducted on an on-prem server of 72 of CPU(Intel(R) Xeon(R) Gold 6140 CPU @ 2.30GHz). The operating system is Ubuntu 18.04.4 LTS. All the codes for the task are written in Python 3.6\cite{ref25}\textit{(\href{https://github.com/Frank-LSY/XGB-SHAP-EHR-EF}{Github Links}$\footnote{https://github.com/Frank-LSY/XGB-SHAP-EHR-EF})$}. The XGBoost model is implemented with xgboost 1.1.1\cite{ref26} and scikit-learn 0.23\cite{ref27}; the SHAP interpretation is implemented with shap\cite{ref28}; t-SNE algorithm is implemented with scikit-learn.manifold.TSNE\cite{ref29}, where the perplexity is set to 100.

\section*{Results}

\begin{wrapfigure}{r}{8cm}
\centering                                            
\subfigure[Regression performance]{                   
\begin{minipage}{5cm}
\centering                                     
\includegraphics[width=\textwidth]{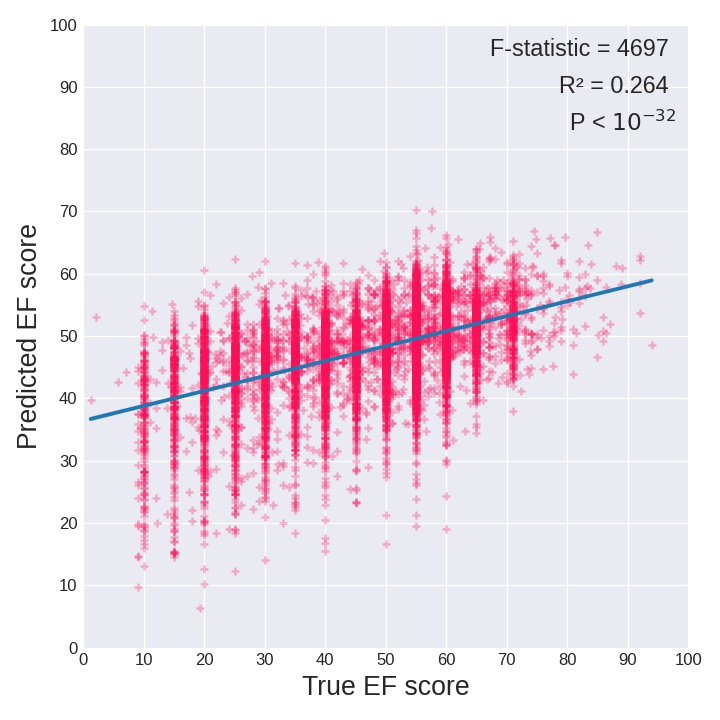}
\end{minipage}}

\subfigure[The first regression tree]{
\begin{minipage}{9cm}
\centering                                           
\includegraphics[width=\textwidth]{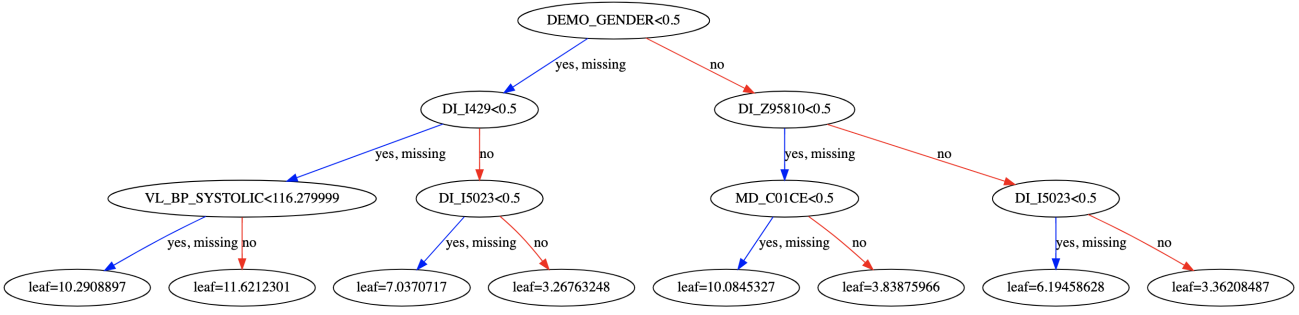}
\end{minipage}}
\caption{}             
\label{fig2}                                        
\end{wrapfigure}

\textit{Predicting performance of XGBoost model}

The performance for XGBoost regression is $RMSE= 12.6303\pm0.00201$ (95\% CI) with 100 random attempts on validation dataset. We evaluated the correlation between the predicted EF  and real EF scores (\textbf{Figure 2A}), of which  $R^2=0.2619$, with a $p < 10^{-32}$. In \textbf{Figure 2B}, we show the first tree (the most prominent out of 100 trees), which illustrate what features are utilized by the tree and how it  splits samples and make final predictions. For example, the first feature for split is the DEMO\_GENDER, which indicate that gender is a very important feature for predicting EF among HF patients. Note that our model only used structured data from EHR to predict a numeric score with moderate accuracy. The predictions clearly follow correct trends and numeric values are moderately accurate. 

\textit{Interpretation with XGBoost and SHAP}

The XGBoost model and SHAP evaluate features from different perspectives. The XGBoost uses the \textit{coverage} to reflect importance of a feature, which denotes the percentage cases in which a feature is utilized during the decision pathway of making the final prediction. On the other hand, SHAP analysis assigns a SHAP value for a feature in each case (i.e., case-specific), which reflects the impact on the feature (measured as deviation from the mean predicted value) when different combinations of features (including or excluding the feature of interest) are used to predict the target value of a case.  The important features provided by two methods are shown  in \textbf{Figure 3}, where \textbf{Figure 3A} is the feature importance generated with XGBoost with coverage$\geq$0.01, and \textbf{Figure 3B} is the top 20 most important features according to   SHAP analysis. %The figure should be interpreted as:

\begin{figure}[h!]
\centering
\includegraphics[width=1.1\textwidth]{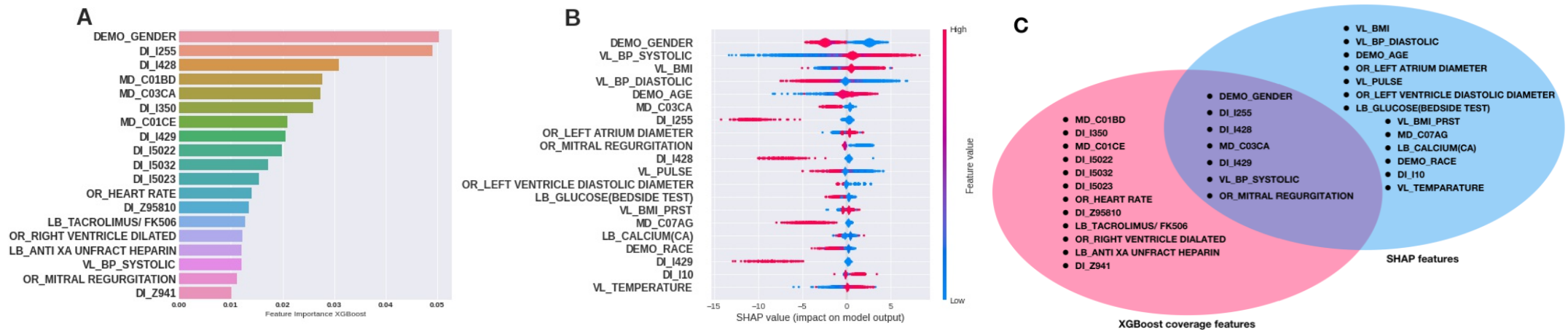}
\caption{}
\label{fig3}
\end{figure}

Interpretation of SHAP scores is as follows. For the SHAP value plot, each row presents the distribution of SHAP values assigned to a feature across all cases. The $x$-axis denote the SHAP value, and the unit reflect how much the presence/absence of a feature with a particular value in a case will lead to deviation of predicted EF score (the unit is  \%) from the mean of prediction values using all possible combinations of feature sets from the case. The pseudo-color of a data point indicates the value of the feature of interest in a case, the further a point deviate the mean of predictions (which is 0), the more impact the features has on the prediction in the case.   Therefore, a positive SHAP value is on the right side of mean on the $x$-axis, indicating the feature with the value in a case leads to target value above the average predicted value, and below the average value on the left side. 
For example, for VL\_BMI, except for extreme cases, the patient's SHAP value is positive when the BMI is higher, and negative when the BMI is lower. That is, for our prediction model, by SHAP interpretation, we get the result that for heart failure patients, they tend to have EF scores higher than average if the BMI value is high and vice versa.

\begin{wrapfigure}{r}{0.5\textwidth}
\centering
\includegraphics[width=0.6\textwidth]{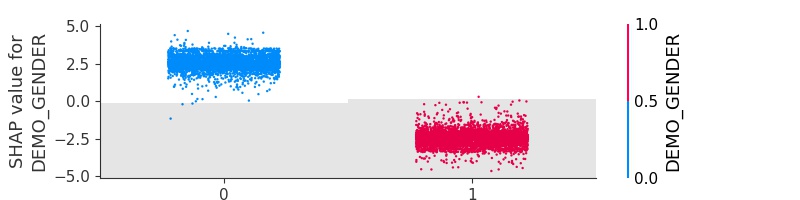}
\caption{}
\label{fig4}
\end{wrapfigure}

As shown in the \textbf{Figure 3C}, the two methods identified a common set of informative features as well as some disjoint features, and the  ranking of features are similar.  First, the most important features that both has highest coverage for XGBoost feature importance and highest SHAP value impact is DEMO\_GENDER. If we dive deeper, like what is shown in \textbf{Figure 4}, we can see that female patients tend to have about 5\% higher EF score compare to male patients (female is represented as 0 and male as 1). Both XGBoost and SHAP interpretations treat DI\_I255 (Ischemic Cardiomyopathy), DI\_I428 (other Cardiomyopathies) and DI\_I429 (Cardiomyopathy, unspecified) as critical diagnoses; MD\_C03CA (Sulfonamides, plain) as critical medical dispenses, OR\_MITRAL REGURGITATION (Mitral valve regurgitation) as critical order results, if presented. Besides, according to  \textbf{Figure 3B}, BP\_SYSTOLIC(systolic blood pressure) and BP\_DIASTOLIC(diastolic blood pressure) all have relatively high SHAP values, but their contributions to the prediction are quite the opposite. As shown in \textbf{Figure 5}, the model tend to assign higher EF score value predictions if patients have higher systolic pressure, while assign lower EF score value predictions if patients have higher diastolic pressure.

\begin{figure}[!h]
\centering
\includegraphics[width=0.8\textwidth]{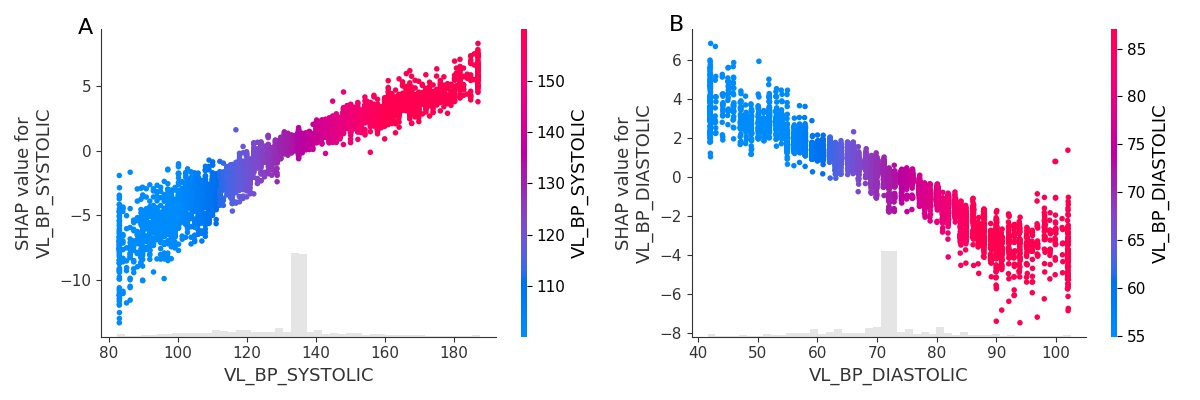}
\caption{}
\label{fig5}
\end{figure}

\textit{Sample clustering analysis with t-SNE and SHAP value}

SHAP analysis provide a new perspective for inspecting data points: case-specific impact of each feature of a data point. This provide us an opportunity to inspect whether different cases share a common pattern (joint distribution) of SHAP scores that characterize a subset of cases, which would provide a new perspective to identify patients share a common underlying disease mechanisms.  We applied the t-SNE algorithm to visualize the distribution of samples in the original feature space as well as in SHAP score space. For the former, each case was represented in the original feature space, and for the latter, each case's SHAP values were used as input features, and t-SNE project the data points from both representation to a 2D space.

\begin{figure}[!h]
\centering 
\subfigure[t-SNE unsupervised feature value clustering for EF score]
{
	\begin{minipage}{7cm}
	\centering
	\includegraphics[width=\textwidth]{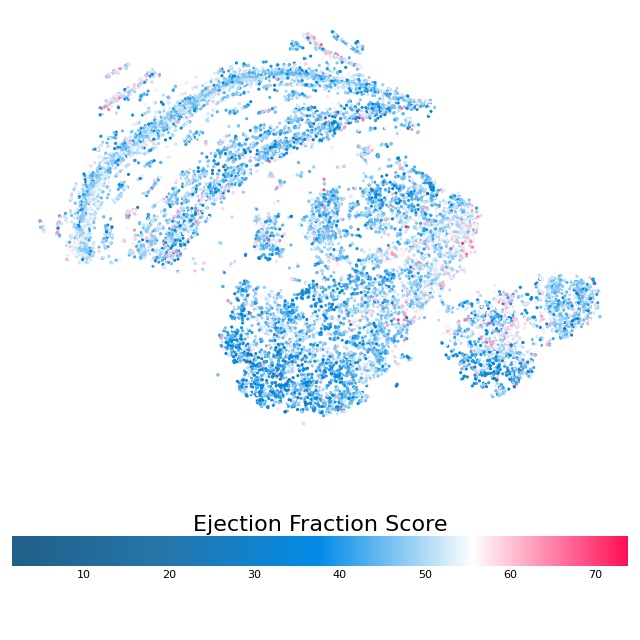}   
	\end{minipage}
}
\subfigure[t-SNE supervised SHAP value clustering for EF score]
{
	\begin{minipage}{7cm}
	\centering
	\includegraphics[width=\textwidth]{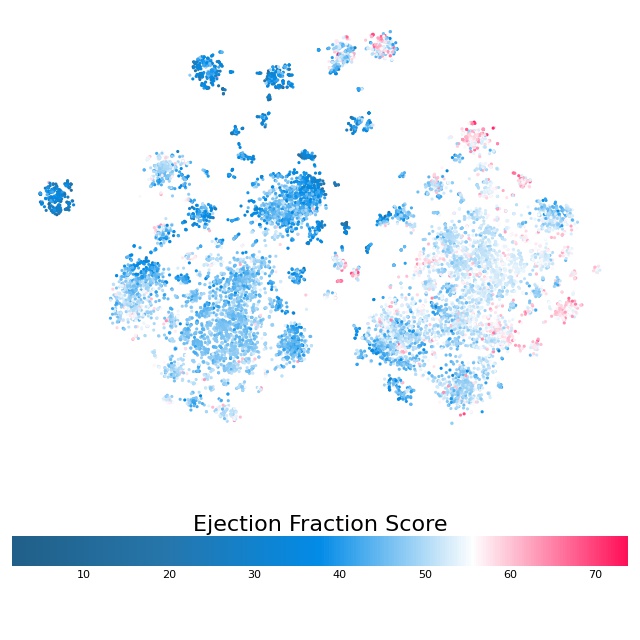}
	\end{minipage}
}
\caption{}
\label{fig6}
\end{figure}

\textbf{Figure 6} shows the general clustering result with original feature values and SHAP values of all features respectively. The color represents EF score for each case. Compared with \textbf{Figure 6A}, which is the clustering of samples based on original feature values, it can be seen that the clustering of SHAP value(\textbf{Figure 6B}) makes the data points with similar EF values closer. More specifically, patients with high EF (HFpEF) appear to be evenly distributed among the patients with reduced EF (HFrEF) in t-SNE results based on the original features, whereas HFpEF patients tend to be more tightly located in when SHAP feature values were used as input. Furthermore, the HFpEF patients form sub-clusters in the SHAP-derived t-SNE space, which indicates that there were distinct combinations of SHAP values (thereby clinical features) that was detectable by the t-SNE algorithm. In summary, representing data points in the SHAP space reveals characteristics of samples that are not detectable in the original data space.   
We then set out to investigate which features contributed to the sub-clusters in the SHAP-derived t-SNE analysis. For the 10 graphs in this section in \textbf{Figure 7}, we plotted each data point in the same position as they were in \textbf{Figure 6B}, and we used pseudo-color to illustrate the original values of different features in each sub-plot. 

\begin{figure}[h!]
\centering
\includegraphics[width=0.9\textwidth]{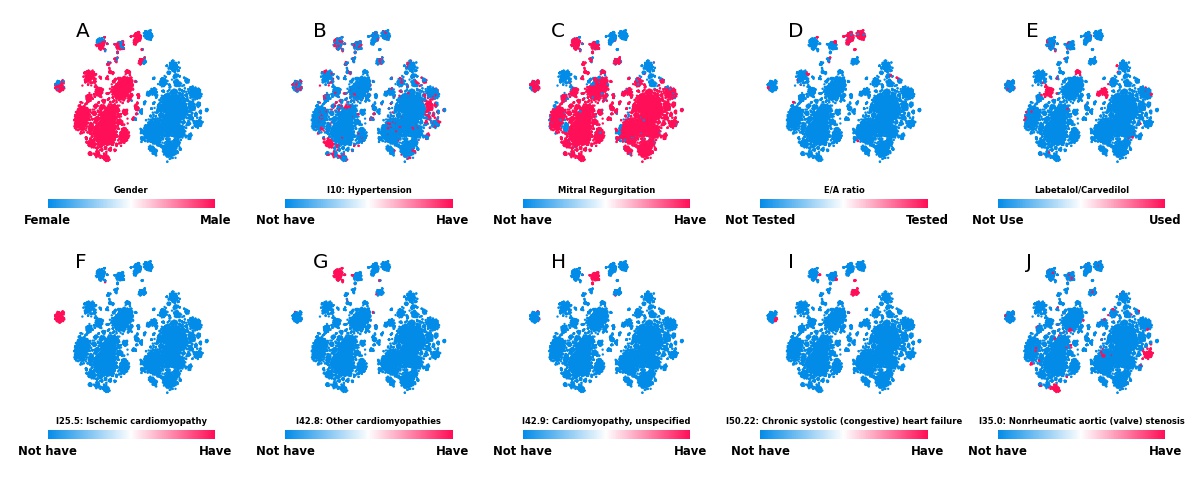}
\caption{t-SNE supervised SHAP value clustering for different features}
\label{fig7}
\end{figure}

When inspecting the combination of features of different sub-clusters, it is interesting to note that gender appears to a major factor that lead to division of two large sub-populations (\textbf{Figure 7A}). This finding agrees with the finding from XGBoost analysis as mentioned previously: gender is the most important feature to be considered when the XGBoost model trying to predict the EF score.  The separation of patients according gender indicate that patients with different gender tend to have distinct combinations of characteristics of other features, which suggest there is a major difference in disease mechanisms of HF patients of differ gender. 
It can also be seen that mitral regurgitation is present in many patients except few smaller clusters. Presence of mitral regurgitation is usually associated with dilated left ventricles, which is usually associated with reduced EF (HFrEF). For the rest of sub-clusters, each has at least one unique feature that clearly assumes a value different from those of majority of the patients, reflecting the importance of such a feature in differentiate sub-populations of HF patients, and in-depth understanding of the characteristics of particular patterns of feature combinations may reveal distinct disease mechanisms underlying these subgroups.   

\section*{Discussion}
\textit{Findings}

In this study, we showed that the XGBoost regression model could be trained to predict EF score (an objective measurement of heart function) with fair performance, only based the structured EHR data. Furthermore, we showed that SHAP analysis revealed information that could not be easily acquired through analyzing original features. The results are encouraging in that they demonstrated the feasibility of monitoring HF patient disease stage, thereby progression of the disease, of HF patients through mining EHR. We anticipate that with more training cases, more importantly more information from detailed clinical notes that provide the information of symptoms and signs of HF patients, the accuracy of the predictive model can be further improved and eventually make such a system clinically applicable. Given the dominant impact of HF on human mortality, a small improvement in monitoring patient disease progression can be translated into a significant improvement in overall patient outcomes. %Besides, with the assist of SHAP interpretation and supervised clustering visualization, we would make the  ML model transparent, and obtain meaningful interpretation of the model and the dataset. The work would further enhance the understanding of heart failure.

Some of our findings are confirmed by previous studies, which indicate the validity of our approach in discovering patterns and important features.  For example, the 5\% difference in EF values between men and women was reported  in several publications. According to Chung et al.\cite{ref30} who reviewed 1435 women and 1183 men cMRI, and concluded whether there is heart failure or not, female have the median EF score of 75\% and male have the median of 70\%, $p<0.001$. This is consistent with a 5\% gender difference of our XGBoost predicted results. Moreover, earlier studies\cite{ref31,ref32,ref33} also reported that in patients with heart failure of different genders, female are more likely to be diagnosed with HFpEF, whereas male are more likely to be diagnosed with HFrEF. Another example is the relation between blood pressure and EF score is another appealing research topic. According to Katsuya et al.\cite{ref34} among patients with acute heart failure syndromes (AHFS), it has been reported that those with a reduced left ventricular ejection fraction (LVEF) tend to be hypotensive or normotensive, whereas those with a preserved LVEF tend to be hypertensive. Their study which evaluated 4831 patients led to the conclusion that patients with an admission SBP$<$120 mmHg were more likely to have a reduced LVEF than a preserved LVEF. In contrast, patients with an admission SBP $\geq$ 120 mmHg were equally likely to have a preserved or reduced LVEF, indicating that there was no relation between a higher admission SBP and the LVEF. However, we did not find references that suggest the changes in systolic and diastolic blood pressure would have opposite effects on EF score. Such combination of characteristics in these features are also discovered in our analyses. Discovery of such combination patterns, thereby sub-population of patients, not only prompts further investigation of potentially distinct underlying disease mechanism but also suggest that tailored prediction/monitoring models should be developed to different sub-populations, aka, a mixture of expert models, to enhance the performance.

\textit{Limitation}

This is a early attempt of using ML models to detect HF patient stage.  Currently, the model only utilize the structured data from EHR, missing significant amount of information from clinical notes. One future direction should be extract informative representation of symptoms and signs associated with HF to enhance the accuracy of prediction.  Another limitation is that current model does not attempt model the temporal trajectory of heart function, which will be addressed in future studies.

\section*{Conclusion}
With XGBoost model, SHAP interpretation and unsupervised clustering visualization, we can 1) predict EF score from tabular EHR data with decent performance; 2)generate interpretation for both the XGBoost model and dataset; 3) classified the subgroups of HF. The generated interpretations is consist with HF diagnose guidelines and humans intuition. Even people with little medical background, after reading the relevant explanations in this article, will have a basic understanding of the risk of heart failure, which is largely related to gender, blood pressure, age, pulse, BMI, some diagnoses (miscellaneous cardiomyopathies), and medications (Sulfonamides, Alpha and beta blocking agents). To a large extent, this indicates that the use of machine learning models to construct clinical decision aids related to heart failure is justifiable and feasible.

\section*{Acknowledgements}
This work is partially support by NIH grant (R01LM012011) to LX.  The authors would like to thank Dr. Zheng Li for discussions. 

\makeatletter
\renewcommand{\@biblabel}[1]{\hfill #1.}
\makeatother

\bibliographystyle{unsrt}

\begin{thebibliography}{1}
\setlength\itemsep{-0.1em}

\bibitem{ref1}
Virani, S. S., Alonso, A., Benjamin, E. J., et al. (2020). Heart disease and stroke statistics—2020 update: A report from the American Heart Association. In Circulation (Vol. 141, Issue 9, pp. E139–E596). Lippincott Williams and Wilkins. https://doi.org/10.1161/CIR.0000000000000757.
\bibitem{ref2}
Weinbrenner, S., Langer, T., Scherer, M., et al. (2012). Nationale VersorgungsLeitlinie Chronische Herzinsuffizienz. In Deutsche Medizinische Wochenschrift (Vol. 137, Issue 5, pp. 219–226). https://doi.org/10.1055/s-0031-1292894
\bibitem{ref3}
Ponikowski, P., Voors, A. A., Anker, S. D., et al. (2016). 2016 ESC Guidelines for the diagnosis and treatment of acute and chronic heart failure. In European Heart Journal (Vol. 37, Issue 27, pp. 2129-2200m). Oxford University Press. https://doi.org/10.1093/eurheartj/ehw128
\bibitem{ref6}
Gutman, R., Shalit, U., Caspi, O., Aronson, D. (2020). What drives success in models predicting heart failure outcome? European Heart Journal, 41(Supplement\_2). https://doi.org/10.1093/ehjci/ehaa946.3556
\bibitem{ref7}
Frizzell, J. D., Liang, L., Schulte, P. J., et al. (2017). Prediction of 30-day all-cause readmissions in patients hospitalized for heart failure: Comparison of machine learning and other statistical approaches. JAMA Cardiology, 2(2), 204–209.
\bibitem{ref8}
Budholiya, K., Shrivastava, S. K., Sharma, V. (2020). An optimized XGBoost based diagnostic system for effective prediction of heart disease. Journal of King Saud University - Computer and Information Sciences. https://doi.org/10.1016/j.jksuci.2020.10.013
\bibitem{ref17}
Chen, T., \& Guestrin, C. (n.d.). XGBoost: A Scalable Tree Boosting System. https://doi.org/10.1145/2939672.2939785
\bibitem{ref4}
Lundberg, S. M., Lee, S. I. (2017). A unified approach to interpreting model predictions. Advances in Neural Information Processing Systems, 2017-Decem(Section 2), 4766–4775.
\bibitem{ref5}
Lundberg, S. M., Erion, G., Chen, H., et al. (2020). From local explanations to global understanding with explainable AI for trees. Nature Machine Intelligence, 2(1), 56–67. https://doi.org/10.1038/s42256-019-0138-9
\bibitem{ref9}
Katuwal, G. J., Chen, R. (2016). Machine learning model interpretability for precision medicine. http://arxiv.org/abs/1610.09045
\bibitem{ref10}
Yang, C., Delcher, C., Shenkman, E., Ranka, S. (n.d.). Predicting 30-day all-cause readmissions from hospital inpatient discharge data.
\bibitem{ref11}
Che, Z., Purushotham, S., Khemani, R., Liu, Y. (2016). Interpretable deep models for ICU outcome prediction. AMIA ... Annual Symposium Proceedings. AMIA Symposium, 2016, 371–380. /pmc/articles/PMC5333206/?report=abstract
\bibitem{ref12}
Luo, G. (2016). Automatically explaining machine learning prediction results: A demonstration on type 2 diabetes risk prediction. Health Information Science and Systems, 4(1), 1–9. https://doi.org/10.1186/s13755-016-0015-4
\bibitem{ref13}
Krause, J., Perer, A., Ng, K. (2016). Interacting with Predictions: Visual Inspection of Black-box Machine Learning Models Data Curation View project Interpreting and Visualizing Machine Learning Models View project Interacting with Predictions: Visual Inspection of Black-box Machine Learning. https://doi.org/10.1145/2858036.2858529
\bibitem{ref16}
apache/lucene-solr: Apache Lucene and Solr open-source search software. (n.d.). Retrieved November 28, 2020, from https://github.com/apache/lucene-solr
\bibitem{ref18}
Xia, Y., Liu, C., Li, Y. Y., \& Liu, N. (2017). A boosted decision tree approach using Bayesian hyper-parameter optimization for credit scoring. Expert Systems with Applications, 78, 225–241.
\bibitem{ref19}
Zieba, M., Tomczak, S. K., \& Tomczak, J. M. (2016). Ensemble boosted trees with synthetic features generation in application to bankruptcy prediction. Expert Systems with Applications, 58, 93–101. https://doi.org/10.1016/j.eswa.2016.04.001
\bibitem{ref20}
sklearn.model\_selection.GridSearchCV — scikit-learn 0.23.2 documentation. (n.d.). Retrieved November 29, 2020, from https://scikit-learn.org/stable/modules/generated/sklearn.model\_selection.GridSearchCV.html
\bibitem{ref21}
Wright, S. J. (n.d.). COORDINATE DESCENT ALGORITHMS FOR LASSO.pdf. 1–27. http://www.optimization-online.org/DB\_FILE/2014/12/4679.pdf
\bibitem{ref22}
Shapley, L. S. (2016). 17. A Value for n-Person Games. In Contributions to the Theory of Games (AM-28), Volume II (pp. 307–318). https://doi.org/10.1515/9781400881970-018
\bibitem{ref23}
API Reference — SHAP latest documentation. (n.d.). Retrieved February 25, 2021, from https://shap.readthedocs.io/en/latest/
\bibitem{ref24}
Hinton, G., \& Roweis, S. (n.d.). Stochastic Neighbor Embedding.
\bibitem{ref25}
Python Release Python 3.6.6 | Python.org. (n.d.). Retrieved November 29, 2020, from https://www.python.org/downloads/release/python-366/
\bibitem{ref26}
dmlc/xgboost: Scalable, Portable and Distributed Gradient Boosting (GBDT, GBRT or GBM) Library, for Python, R, Java, Scala, C++ and more. Runs on single machine, Hadoop, Spark, Dask, Flink and DataFlow. (n.d.). Retrieved November 29, 2020, from https://github.com/dmlc/xgboost/
\bibitem{ref27}
Release Highlights for scikit-learn 0.23 — scikit-learn 0.23.2 documentation. (n.d.). Retrieved November 29, 2020, from https://scikit-learn.org/stable/auto\_examples/release\_highlights/plot\_release\_highlights\_0\_23\_0.html
\bibitem{ref28}
slundberg/shap: A game theoretic approach to explain the output of any machine learning model. (n.d.). Retrieved November 29, 2020, from https://github.com/slundberg/shap
\bibitem{ref29}
sklearn.manifold.TSNE — scikit-learn 0.23.2 documentation. (n.d.). Retrieved November 29, 2020, from https://scikit-learn.org/stable/modules/generated/sklearn.manifold.TSNE.html
\bibitem{ref30}
Chung, A. K., Das, S. R., Leonard, D., et al. (2006). Women have higher left ventricular ejection fractions than men independent of differences in left ventricular volume: The Dallas heart study. Circulation, 113(12), 1597–1604. https://doi.org/10.1161/CIRCULATIONAHA.105.574400
\bibitem{ref31}
Savarese, G., \& D’Amario, D. (2018). Sex differences in heart failure. In Advances in Experimental Medicine and Biology (Vol. 1065, pp. 529–544). Springer New York LLC. https://doi.org/10.1007/978-3-319-77932-4\_32
\bibitem{ref32}
Duca, F., Zotter-Tufaro, C., Kammerlander, A. A., Aschauer, S., Binder, C., Mascherbauer, J., \& Bonderman, D. (2018). Gender-related differences in heart failure with preserved ejection fraction. Scientific Reports, 8(1), 1080. https://doi.org/10.1038/s41598-018-19507-7
\bibitem{ref33}
Regitz-Zagrosek, V. (2020). Sex and Gender Differences in Heart Failure. International Journal of Heart Failure, 2(3), 157. https://doi.org/10.36628/ijhf.2020.0004
\bibitem{ref34}
K, K., N, S., Y, S., \& T, T. (2013). Relationship between systolic blood pressure and preserved or reduced ejection fraction at admission in patients hospitalized for acute heart failure syndromes. International Journal of Cardiology, 168(5). https://doi.org/10.1016/J.IJCARD.2013.07.226
\end{thebibliography}

\end{document}